\documentclass[10pt,twocolumn,letterpaper]{article}

\usepackage{iccv}
\usepackage{times}
\usepackage{epsfig}
\usepackage{graphicx}
\usepackage{amsmath}
\usepackage{amssymb}
\usepackage{authblk}
\usepackage[greek,english]{babel}

\usepackage[breaklinks=true,bookmarks=false]{hyperref}

\iccvfinalcopy % *** Uncomment this line for the final submission

\def\iccvPaperID{****} % *** Enter the ICCV Paper ID here

\ificcvfinal\fi

\begin{document}

%%%%%%%%% TITLE
\title{A New Dataset for End-to-End Sign Language Translation: The Greek Elementary School  Dataset}
\newcommand*\samethanks[1][\value{footnote}]{\footnotemark[#1]}
\author[1]{Andreas Voskou\thanks{ ai.voskou@edu.cut.ac.cy} }
\author[1]{Konstantinos P. Panousis}
\author[2]{Harris Partaourides}
\author[1]{\\Kyriakos Tolias}
\author[1]{Sotirios Chatzis}

{
    \makeatletter
    \renewcommand\AB@affilsepx{: \protect\Affilfont}
    \makeatother

    \makeatletter
    \renewcommand\AB@affilsepx{, \protect\Affilfont}
    \makeatother

\affil[1]{Cyprus University of Technology}
\affil[2]{AI Cyprus Ethical Novelties Ltd}

}
\maketitle
\ificcvfinal\thispagestyle{empty}\fi

%%%%%%%%% ABSTRACT
\begin{abstract}
Automatic Sign Language Translation (SLT) is a research avenue of great societal impact. End-to-End SLT  facilitates the interaction of Hard-of-Hearing (HoH) with hearing people, thus improving their social life and opportunities for participation in social life. However, research within this frame of reference is still in its infancy, and current resources are particularly limited. Existing SLT methods are either of low translation ability or are trained and evaluated on datasets of restricted vocabulary and questionable real-world value. A characteristic example is Phoenix2014T benchmark dataset, which only covers weather forecasts in German Sign Language. To address this shortage of resources, we introduce a newly constructed collection of 29653 Greek Sign Language video-translation pairs which is based on the official syllabus of Greek Elementary School. Our dataset covers a wide range of subjects. We use this novel dataset to train recent state-of-the-art Transformer-based methods widely used in SLT research. Our results demonstrate the potential of our introduced dataset to advance SLT research by offering a favourable balance between usability and real-world value. 
\end{abstract}

%%%%%%%%% BODY TEXT
\section{Introduction}
\label{sec:intro}

Sign Language (SL) is a medium of communication that primarily uses hand gestures, facial expressions, and body movement to convey a speaker's thoughts,  forming  a  complete and formal language. It is the primary means of communication for deaf individuals. National Sign Languages, being the native languages of deaf SL users, are a vital aspect of cultural diversity in Europe and the world. Access to SL communication is essential for HoH as it enables access to equal education, employment, and healthcare services. In Europe, there are 30 official Sign Languages and over 750,000  SL users, but only 12,000 interpreters. This shortage undermines the right to equal education and health  and often endangers the lives of deaf people.

In contrast to the common misconception, Sign Languages are completely independent natural languages. Each national SL is unique with its own grammar, syntax, and vocabulary. Additionally,  there is no direct connection between a spoken language and its corresponding Sign Language, for example, Greek and Greek Sign Language.  A true  Sign Language Translation (SLT) system  needs to capture the visual patterns in the signed signal, decompose their linguistic meanings, and reconstruct them into spoken language text. These facts make the SLT task particularly challenging from a technical point of view.

Despite the importance of SLT systems, progress in this field has been limited. The current research has almost exclusively focused on simpler tasks, such as the recognition of static or dynamic gestures. Furthermore, the vast majority of successful SLT models currently available are trained on the Phoenix2014T \cite{Camgoz18}, which is a single-topic dataset with a limited vocabulary, or similarly restricted datasets such as CSL-Daily \cite{zhou2021improving}. While models trained on more meaningful datasets do exist, usually they either yield outcomes of significantly low quality or suffer from other types of limitations. These facts make it clear that there is a need for better machine learning techniques and, consequently, a need for more and better training datasets.

This work addresses this problem by introducing a new SLT-suitable dataset, the Greek Elementary School Dataset (Elementary23). This dataset comprises more than 28,000 videos of Greek SL, totaling over 70 hours, each paired with its corresponding spoken Greek translation in text. Prioritizing optimal technical quality, the data was captured using high-definition cameras and featured expert signers. All examples are derived from the teaching materials of Greek elementary schools, covering a broad spectrum of subjects.

\section{Related Work}

\paragraph{Datasets.} 

Sign Language Processing involves a variety of tasks; the most common and well-studied  is Sign Language Recognition (SLR) \cite{al2022new,zuo2022c2slr,bohavcek2022sign}. SLR is technically  a special case of video classification,  where a (short) video sequence is assigned a single label. Datasets designed for the particular task include  SL videos  annotated  with the so-called Glosses; these are text-like labels that express discrete SL expressions. Glosses should not be mistaken for text, since they do  not form a proper or complete language  format and often lack expressive power. Representative  SLR dataset  are the DGS Kinect 40 dataset \cite{Ong_Sign_2012, Elliott_Search_2011} in German Sign Language, the GSL  dataset \cite{adaloglou2020comprehensive} in Greek SL   and many more \cite{Ronchetti2016,li2020word,agris,Cooper2017,asllvd,chai2015devisign,zhang_gesture,wang_gesture,ozdemir2020bosphorussign22k,camgoz2016bosphorussign}.

The availability  of SL-related datasets is clearly considerable. However, only a select few are suitable for the most critical application of SL processing, namely end-to-end SLT. A dataset is suitable for SLT model training if it possesses two crucial characteristics: i) it includes SL videos paired with corresponding translations in a formal spoken language, and ii) the video-text pairs comprise complete and syntactically correct sentences/phrases of adequate length.

In this context, Phoenix2014T \cite{Camgoz18} dataset has become one of the most extensively studied datasets for this purpose. It features weather news performed in German Sign Language,  and includes both Gloss annotation and text translations. The inclusion of Gloss annotation, combined with its dense single-topic vocabulary, renders the Phoenix2014T dataset more conducive to deep learning and has thus attracted significant attention. 

Other notable datasets in this domain include SWISSTXT-NEWS \cite{camgoz2021content4all} and VRT-NEWS \cite{camgoz2021content4all}, which cover a broader range of topics. These datasets are composed of  {TV}-news data in German and Flemish Spoken and Sign Languages. However, the  results of any end-to-end  SLT attempts on these datasets have been disappointingly low, as they have failed to achieve even remotely acceptable translation quality.  A recent important addition to this field was a project published by the BBC \cite{albanie2021bbc}, consisting of  a particularly large number of  phrases in British Sign Language and English spoken language. The potential  of this new  dataset is especially high, mainly due to its extensive size. Nonetheless, end-to-end SLT results of deep learning models trained on this dataset have yet to appear in the related literature. Further  examples are   the CSL-Daily on Chinese SL with properties similar to  Phoenix2014T and the American-SL dataset How2Sign \cite{duarte2021how2sign} with good  size and quality but lower reported results. A very  recent and  important addition is the  OpenASL dataset \cite{shi2022open}, published in late 2022, covering 300h of American SL videos collected from online videos and demonstrating respectable results. 

\paragraph{Sign Language Translation.} Given the societal importance  SLT, the field can be considered clearly underexplored. However, in recent years, a noteworthy increase of effort has taken place. The 2018 paper \cite{Camgoz18} has been the seminal work on the Phoenix2014T dataset. It implemented recurrent seq-to-seq architectures for modelling, which resulted in promising BLEU-4 scores; these ranged from 10 up to 19 for  different SLT variants. In 2020, the authors of \cite{Camgoz20}  utilized the power of  Transformer networks   in the form of  the Sign Language Transformer and achieved major improvements in the translations. The approach used an S2T architecture and a  feature-extracting  element pre-trained as part of an SLR engine. Using the mentioned setups and  additional Gloss-level supervision,  they  achieved  clearly superior results.  The 2021 paper \cite{Voskou_2021_ICCV} proposed a breakthrough variant of Transformers, which uses a novel form of activation functions which yield \emph{sparse} and \emph{stochastic} representations. This is achieved via a local stochastic competition mechanism that gives rise to stochastic local winner-takes-all (LWTA) units. The method managed to improve the translation quality even more without  auxiliary Gloss supervision. In addition, they showed that  the proposed method can be tuned to reduce the post-training memory footprint by properly exploiting  model uncertainty. Other similar works include \cite{camgoz2020multi}, based on a multistream Transformer,  and approaches like \cite{jin2022prior} and \cite{xie2021pisltrc}  that leverage supplemental data through  transfer learning and  augmentation techniques to gain some additional improvement. 
\section{The {Elementary23} Dataset}
\label{sec:intro}

\subsection{Core Elements}

The introduced  Greek Elementary School dataset \footnote{https://zenodo.org/record/7847052}, dubbed Elementary23, constitutes a noteworthy contribution to the existing body of literature due to its exceptional quality and the substantial number of examples included. Table \ref{tab:elem_statistics} presents a comparative analysis of Elementary23 and the widely utilized Phoenix2014T dataset. As we show, Elementary23 exhibits superior technical and linguistic characteristics.

\begin{table}[h!]
    \centering
        \caption{Elementary23   statistics vs  Phoenix2014T.}
    \begin{tabular}{|l|c c|}
    \hline 
        & Phoenix2014T & Elementary23\\\hline  \hline
        Signers & $9$ & $9$\\\hline
        Total  Hours & $25$ & $71$ \\ \hline
        Total Frames  & $\approx 1.1$M & $\approx 6,3$M\\\hline
        Sentences &  $8257$ & $29653$\\\hline % 6841  29692 
        Vocabulary & $2887$ & $23204$\\\hline %2048  27122
        Singletons & $1077$ & $10126$ \\ \hline %644  13056
        Resolution & $210 \times 260$ & $1280 \times 720$  \\ \hline
        FPS & $25$ & $25$\\\hline
    \end{tabular}
    \label{tab:elem_statistics}
\end{table}

\subsection{Collection Procedure}

The Elementary23 dataset stands in contrast to many relevant datasets as it was not built by annotating preexisting sign language (SL) videos from online or other sources.  Instead, it was assembled through the recording of sign language interpretations of authentic elementary school content. Furthermore the final content was rigorously curated and selected by expert staff, ensuring its high impact and practical value to the deaf community and students.

\begin{figure}[ht]
    \centering
    \includegraphics[scale=0.115]{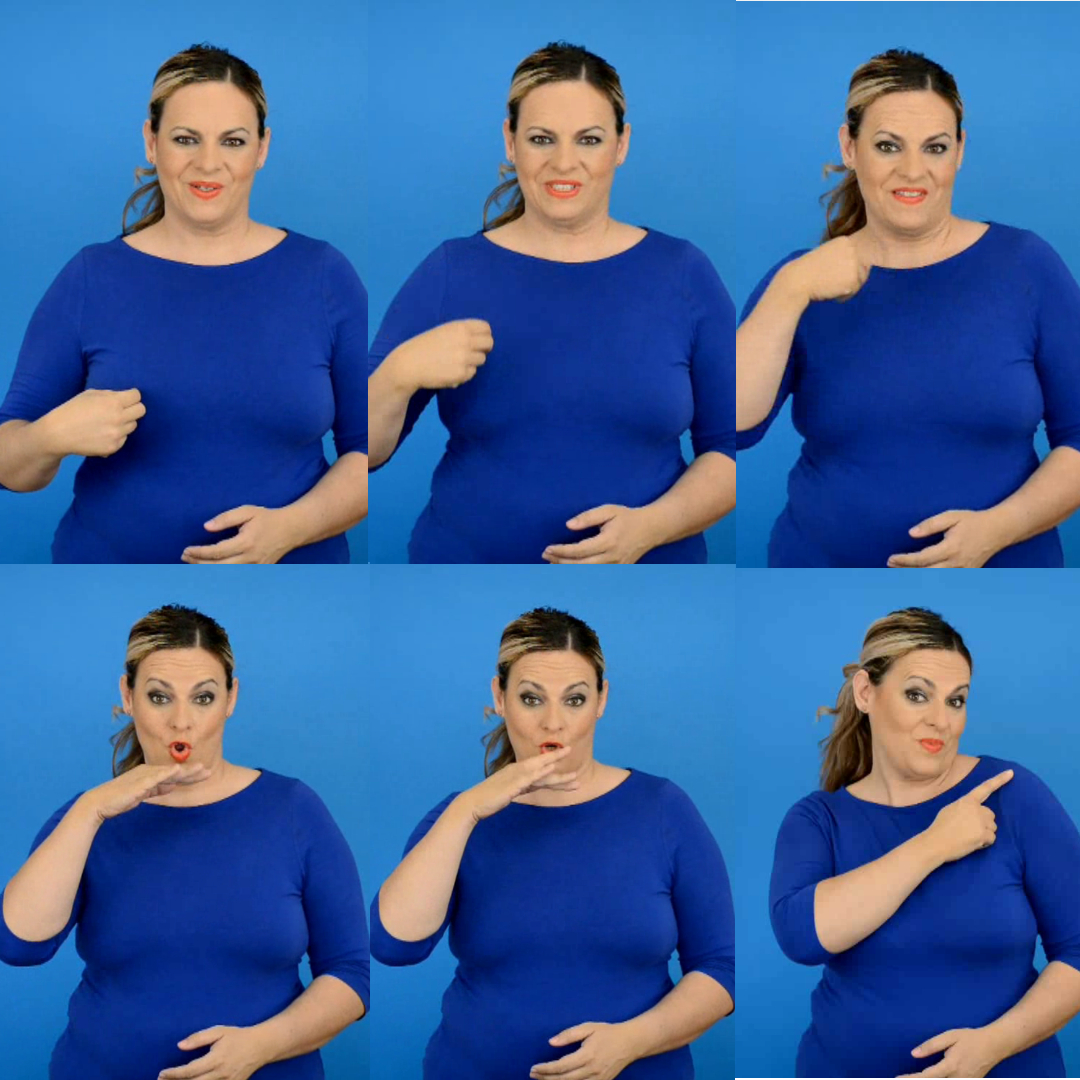} 
      \caption{ Example of Elementary23 Sign Language Video Frames }
    \label{fig:dataexample}
\end{figure}

The recordings for the Elementary23 dataset were made in an environment optimally suited for the task. This included a fixed single-color background and ideal lighting conditions (see Fig. \ref{fig:dataexample}). Professional-grade cameras and equipment were used to record videos at 720p resolution and a frame rate of 25fps.

\begin{figure}[ht]
    \centering
    \includegraphics[scale=0.35]{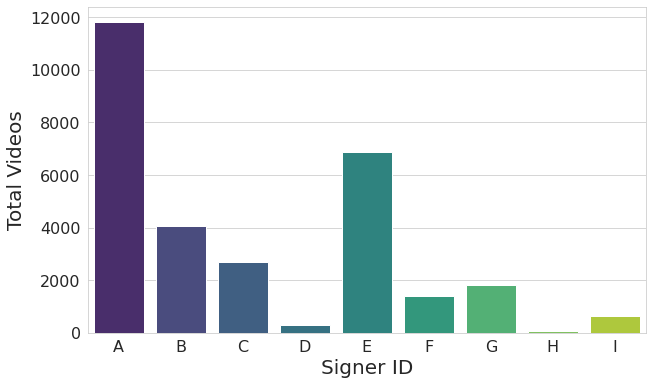} 
      \caption{ Distribution of video-translation pairs per signer }%
    \label{fig:persigner}%
\end{figure}

The material was organised into sentences/phrases and then assigned to nine signers,  all proficient users of Greek Sign Language with extensive knowledge and experience. As a result, the dataset is of exceptional quality, with minimum kinesthesiological and technical errors. The distribution per signer is illustrated in Figure \ref{fig:persigner}.

\subsection{Content and Vocabulary}
 
A Notable aspect of Elementary23 is its broad thematic spectrum. As previously mentioned the dataset is based on  the official syllabus of Greek elementary schools, including  the subjects of Greek Language, Mathematics, Religion Study, Environmental Study, History, and Anthology. The combination of those subjects ensures  a sizeable lexicon of 23,204 words; yet, it is important to note that each individual subject  contributes broad content and an  extensive vocabulary. The statistics for each subject are presented in Table \ref{table:elementary_domain}. The largest in terms of video-translation pairs quantity is the subject of the "Greek Language", which contains 9499  examples; the smallest one is  "Religion Study", with 1,825 entries. Vocabulary-wise, the most comprehensive subject is Anthology which includes a total of 14,741 different words; on the other end, "Mathematics" have the smallest vocabulary with 6,457 words.

\begin{table}[ht]   
    \centering
        \caption{ Number of examples per Subject and  vocabulary metrics}

    \begin{tabular}{|c|cc|}
    \hline
         &  Vocabulary &    Examples \\ \hline\hline %signletons
        Anthology  & 14741   & 4158 \\ \hline  % 10148
        Greek Language &  14345   & 9499 \\ \hline %8.783
        Mathematics &  6457   & 6583 \\ \hline % 3.164
        History &  7716  & 2067 \\ \hline % 5000
        Envir. Study   &  9489   & 5521 \\ \hline % 5719
        Relig. Study  &  8087  & 1825 \\ \hline % 5588
       
    \end{tabular}
    \label{table:elementary_domain}
\end{table}

%------------------------------------------------

		%------------------------------------------------
\begin{table*}
\begin{center}
\caption{    Elementary23-SLT vs key Benchmarks  }

\begin{tabular}{|r||c|ccc|}
\hline
                & Elementary23-SLT     & Phoenix2014T \cite{Camgoz18} & SWISSTXT-NEWS \cite{camgoz2021content4all} & VRT-NEWS \cite{camgoz2021content4all}                    \\

 \hline
\hline

 Sentences & 8372 & 8257 & 6031  & 7174 \\
 
Total Words            &     83327                        &      99081                                          &         72892                                       &          79833                                \\

Vocabulary  &                                 8202                     & 2887                                              & 10561                                              & 6875                                          \\

Mean Word Freq.                &                          10.16  &      34.3                                          &         6.9                                       &          11.6                                \\

Sigletons                    &         3327 ($41\%$)                  & 1077              ($37\%$)                                & 5969  ($57\%$)                                             & 3405           ($50\%$)                               \\

Rare words (\textless{}5)              &          6155  ($75\%$)             & 1758   ($60\%$)                                           & 8779             ($83\%$)                                  & 5334                 ($78\%$)                         \\

\hline
\end{tabular}
\label{table:det_stats}
\end{center}

\end{table*}

\subsection{SLT Subset}

The main motivation of Elementary23 is to contribute one of the largest Sign Language datasets paying particular emphasis to technical and linguistic excellence. However, the dataset is not necessarily an ideal candidate for training end-to-end SLT deep networks. Specifically, there are aspects of Elementary23 that present significant modelling challenges for deep networks. These include the dataset's high word sparsity; the  high number of singletons (words appearing only once in the corpus); the inclusion of particularly small phrases; and the limited number of frequently-appearing words. To be fair, this is not a problem with the dataset itself but rather a limitation of modern deep networks, which typically require multiple examples to learn from data. To overcome this issue, in this work we also present an appropriate representative subsample of the dataset, which we dub Elementary23-SLT \footnote{\href{https://figshare.com/articles/dataset/Elementary23_SLT/22262953}{https://figshare.com/articles/dataset/Elementary23\_SLT/22262953}}. This is more suitable for training end-to-end SLT deep networks, and has  a size similar to Phoenix2014T and other recently published benchmark datasets.

The selection process was guided by three primary principles: (i) decrease the absolute number of singletons; (ii) increase the density of frequent words and bigrams; and (iii) keep  content diverse. To achieve these targets, we first went through preliminary cleaning, whereby we eliminated  singleton-only sentences. Afterwards, we ran a simple  dynamic multi-round elimination process:
At each round, we listed all sentences containing singletons but no frequent elements, and we eliminated approximately a quarter of them. We then recalculated word frequencies and redefined singletons and frequent words based on the surviving subset. At the end of each round, we confirmed that all six subjects were represented with a sufficient number of remaining examples, at least $10\%$ of the original.  We repeated this procedure several times, until we ended up with a sample size comparable to the standard benchmarks. 

The aforementioned process left us with a sample of 7168 sentences. Subsequently, we split the data into the typical train, validation and test subsets. During the selection of the validation and test sets, we tried to avoid sentences shorter than four words long, and made sure to exclude all the sentences appearing twice by the same speaker. As an extra processing step, we augmented the training set with an additional 1,204 non-singleton single-word videos. Finally, for comparison reasons, we additionally performed a train-validation-test split on the entire  dataset following similar principles regarding duplicated entries. We will be referring to this split as Elementary23-Raw. In both cases, all speakers may participate in all the subsets.
 
The final SLT set contains $8372$ video-sentence pairs, organized as $7348$ pairs for the training set, $512$ for the validation set, and $512$ for the test set. Through this operation, we produced an effective subset in the typical size spectrum, that retains the desired qualitative elements of the complete collection while exhibiting some improved quantitative factors. Key statistics regarding the  subset and benchmarks are stated in Table \ref{table:det_stats} and will be analysed  later in this Section.

To enable the examination and exploitation of the data by the research community, we have ensured that Elementary23-SLT  complies with established standards regarding size and structure. Additionally,  the data will be available in a file format that is practical and consistent with prior works \cite{Camgoz20, Voskou_2021_ICCV}. Specifically, we have created a JSON file comprising a list of dictionaries, each corresponding to a particular video-sentence pair. These dictionaries encompass all auxiliary elements, such as numbering and signer ID, as well as the principal input-output data; the latter consists of the frame-wise feature sequence and the corresponding translation in modern Greek.

While  feature extraction  is technically a component of  deep network  development, we have adopted conventional practice by embedding the extracted features within the introduced SLT dataset. This approach does not affect the deep network training process or the final product, yet it considerably reduces the costs and effort of model development. The complete videos will also be made available.

\subsection{Landmarks and  Trajectories}

State-of-the-art SLT networks often utilize convolutional subparts as feature extractors; these are pre-trained on sign language recognition datasets. Such subparts can extract spatial information from video frames by leveraging their prior knowledge of core sign language elements. However, this preprocessing
phase requires laborious dataset annotation in terms of auxiliary Glosses. Therefore, this process is of reduced applicability: developed models can only generalize on other datasets of similar Gloss structure. In addition, it is  incompatible with the Greek Elementary School dataset: its immense size renders provision of Glosses completely out of scope.

\begin{figure}[ht]
    \centering    \includegraphics[scale=1.2]{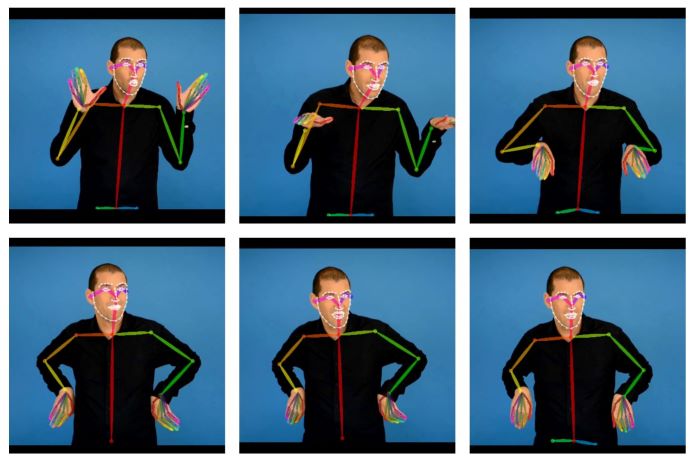} 
      \caption{ Body Landmarks - Trajectories  }%
    \label{fig:openpose}%
\end{figure}

To address this issue, we follow an alternative, yet established approach that involves using the OpenPose engine \cite{cao2017realtime}. The OpenPose engine is a convolutional neural network that has been trained to track and extract the trajectories of key human body parts. We use OpenPose to track  landmarks related to the 2D positioning of upper body movements, facial expressions, and hand shapes; one can use these as input to a developed end-to-end SLT model. We further scrutinize the extracted features, by excluding components that appear not to contribute enough or exhibit persistently low volatility, such as lower body landmarks. The resulting vector effectively summarizes the video frames and serves as a sufficient source of information for subsequent network layers. Figure \ref{fig:openpose} illustrates a representative example of this approach.

\subsection{Lexical  Statistics  and  Benchmarks  }

While Phoenix2014T has gained popularity as a standard benchmark for SLT due to its ease of modelling, its appropriateness as a benchmark for comparing to the newly introduced dataset remains questionable. This is due to the following facts: i) it covers only a single topic,  in contrast to the multi-subject nature of Elementary23; ii)  vocabulary coverage is very limited; iii)  it includes many sentences of similar structure and content, due to the weather forecasting's strict format; and iv) Phoenix2014T  includes auxiliary Gloss annotation.   

To address these challenges, we're supplementing our benchmark schema with two more datasets: SWISSTXT-NEWS and VRT-NEWS. Containing 6031 and 7174 videos/sentences respectively, they're based on HD TV news videos. We chose them due to their similar size to Phoenix2014T and Elementary23-SLT, good video quality  and their basis in European sign languages, ensuring  directly comparable grammatical structures.

 In Table \ref{table:det_stats}, we present a detailed comparison of Elementary23-SLT and the three selected benchmark datasets on the grounds of various vocabulary-oriented metrics. These metrics were critical in determining the suitability of SWISSTXT-NEWS and VRT-NEWS as the primary benchmarks, since their statistics closely resemble those of Elementary23-SLT. Specifically, both SWISSTXT-NEWS and VRT-NEWS contain similarly sized vocabularies, with 10561 and 6875 sentences, respectively. Thus, they deviate no more than 25 $\%$ from our proposed subset of 8202 sentences.  Furthermore, the percentage of rare words, defined as words that appear less than five times, ranges from 75 $\%$ to 83$\%$  for all three datasets. Finally, the mean word frequencies are comparable across these datasets, with the words in SWISSTXT-NEWS appearing an average of 11.6 times, 6.9 for VRT-NEWS, and 10.1 for Elementary23-SLT.

Notably, Phoenix2014T has a vastly reduced vocabulary size compared to the rest of the considered datasets, which constitutes a central constraint. Additionally, Phoenix2014T is characterized by a considerably higher mean word frequency equal to 34.3. This number is rooted in the limited vocabulary and allows for easier training since it narrows down  verbal varieties. Furthermore, Phoenix2014T has the lowest percentage of rare words and singletons. In terms of singletons, Elementary23-SLT is a close second, with an increase of only 4$\%$.

% selection algo
\section{Translation Methodology}
\label{sec:intro}

In order to tackle the challenging task of end-to-end SLT, we adhere to the guidance of recent developments in the field that advocate for the use of Transformer-based architectures \cite{Camgoz20,Voskou_2021_ICCV,yin2020}. Specifically, we employ the seminal SLT model of \cite{Camgoz20}, and the later  sLWTA-Transformer \cite{Voskou_2021_ICCV} variant; we focus more on the latter method, due to its many advantages. These techniques have been demonstrated to be effective on the well-known PHOENIX-14T dataset, and achieve state-of-the-art results with BLEU-4 scores in the area of 22 and  24, respectively.
\begin{figure}[ht]
    \centering    \includegraphics[scale=0.41]{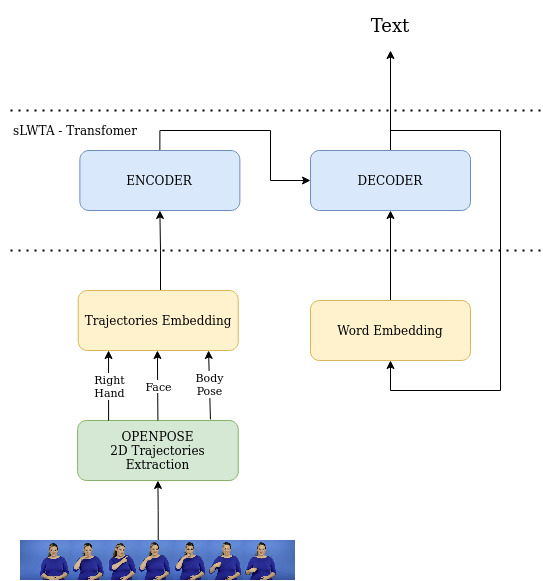} 
      \caption{ The suggested Sign to Text Transformer Network }%
    \label{fig:fullmodel}%
\end{figure}

Transformers are modern deep architectures that rely purely on the Attention mechanism to process temporal dynamics in sequential observations. A standard Transformer layer includes a self-attention layer,  paired with an immediately succeeding Relu-activated Feed-Forward Network (FFN). Sign language Transformers comprise encoder and decoder parts similar to \cite{Vaswani2017}. The encoder is presented with  frame-wise feature vectors obtained from a spatial feature extractor; it learns to process their interactions over the temporal axis and yields  action-aware representations. The decoder uses the latter and  re-expresses the  meaning into formal spoken language. 

The sLWTA SL-Tranformer reapproaches the standard architecture by introducing two forms of stochasticity:  (i) Gaussian weights with posterior distributions estimated through variational inference, instead of standard point estimators on the weights; and (ii) the stochastic local-winner-takes-all layer as a more sophisticated  FFN.

In more detail, we consider a  Bayesian treatment of the weights by training a Gaussian variational posterior; this encloses an uncertainty estimation of each weight,  formed as  $q(w)  = N(\mu,\sigma^2)   $  where $\mu$, $\sigma^2$  are the posterior mean and variance.  For inference,  weight values  are sampled from the Gaussian posteriors in a Monte Carlo fashion. The  reparameterization trick of \cite{kingma2013auto} is employed to  allow for gradient descent-based training. 

The stochastic local-winner-takes-all layer \cite{panousis2019nonparametric}, reported as LWTA or sLWTA, is a sophisticated   non-linear layer that replaces the usual relu-activated dense layers with notable success \cite{panousis2021,panousisnips, kalais2022stochastic,panousis2022competing}.    Let  $\boldsymbol{x} \in R^J$  be the input of a typical dense layer, and $\boldsymbol{y} \in R^H$ the corresponding output vector gained through  multiplication with a weight matrix $W\in R^{J\times H}$ and  activation via a nonlinear function such as ReLU $\boldsymbol{y} = ReLU (W\boldsymbol{x})$ .  LWTA works by  organising  the output $\boldsymbol{y}$ into $K$ blocks of $U$ members/competitors  each,  and  the weight matrix $W$ into K respective  submatrices. For any block indicated by $ k \in \{1,2,..K\} $, we denote as $\boldsymbol{y}_k \in R^U$ and $W_k\in R^{J \times U}$ the corresponding  subparts of $y$ and $W $.  The members of each block compete with each other, and only one of them, the winner, gets activated given an input, while the rest are set to 0.  The so-called  competition is an inter-block stochastic procedure based on sampling the winner from a  discrete  posterior with logits proportional to the linear computation in each unit: 
\begin{equation}
\label{eqn:dense_xi}
q(\boldsymbol{\xi}_k) =  \mathrm{D}\bigg(\boldsymbol{\xi}_k \bigg| \mathrm{softmax}\big( W_k \boldsymbol{x}   ) \bigg),
\;  \; \forall  k \in \{1,2,..K\}
\end{equation}
where $\boldsymbol{\xi}_k \in onehot(U) $   are discrete latent one-hot vectors indicating the winner of each block. 

The sampling process is  effectively approximated  using the Gumbel-Softmax relaxation trick \cite{jang2017categorical} . Controlled  by a temperature hyper-parameter $T$, this technique can offer low-variance gradients during the training phase (high $T$), and hard  discrete  samples,  almost identical to Eq. (\ref{eqn:dense_xi}),  during inference (low $T$).

Finaly, using the postulated $\boldsymbol{\xi}$ latent variables, layer output $y$ can be expressed as follows:
\begin{align}
\boldsymbol{y}_{k} =     \boldsymbol{\xi}_{k}\odot ( {W}_k \boldsymbol{x}_k ), \;  \; \forall  k \in \{1,2,..K\}
\end{align}

where $\odot$ stands for element-wise multiplication.

\subsection{Training and Inference}

In the following, we will be referring to the standard Transformer-based SLT model of \cite{Camgoz20} as the deterministic model. On the other hand, the variant of \cite{Voskou_2021_ICCV} will be dubbed as the stochastic model. 

The training objective of the deterministic SLT model will be to minimise the cross-entropy error between each predicted word and the corresponding label, under a standard seq-to-seq rationale.  When it comes to   inference using the trained model, we again  follow  the usual practice and  run autoregressive decoding, where words of each produced sentence are predicted one by one, and the decoder is presented the encoded representations and the previous predictions. Additionally, we use the beam search algorithm, the parameters of which are optimised on the validation set.  

 Training of the  stochastic variant is slightly more complex. The optimization objective is the negative evidence lower bound (ELBO) of the model, computation of which requires prior assumptions regarding the distributions of the winner indicator latent variables, $\boldsymbol{\xi}$ on each LWTA layer, as well as the trainable weights, $\boldsymbol{w}$, throughout the network. For convenience, we postulate a priori spherical Gaussian weights of the form $p(\boldsymbol{w}) = \mathcal{N}(\boldsymbol{0}, \boldsymbol{I})$, and a symmetric Discrete prior over the winners: $p(\boldsymbol{\xi})  = \mathrm{Discrete}(1/U)$.

Then, the training objective comprises: (i) the standard cross-entropy of the network, with the expressions of the latent variables, $\boldsymbol{\xi}$, expressed via the Gumbel-softmax reparameterization trick, and the Gaussian weights, $\boldsymbol{w} \sim q(\boldsymbol{w})$, expressed via the standard reparameterization trick for Gaussians; (ii) the Kullback-Leibler divergences between the posterior and the prior of the latent variables, $\boldsymbol{\xi}$, and the  Gaussian weights, $\boldsymbol{w}$ \cite{Voskou_2021_ICCV}.

For inference, we directly draw weight samples $\boldsymbol{w}$ from the trained Gaussian posteriors. Similarly, we directly draw samples of the  winner-indicating latent vectors, $\boldsymbol{\xi}$, from the related discrete posteriors. The final prediction is obtained through Bayesian averaging; we sample the weights and $\boldsymbol{\xi}$ parameters from the posteriors four times, calculate the final logits, and average the results. This way we obtain the final logits that we use to drive  beam search, similar to the deterministic model.

\section{Experimental Results}

\begin{table*}[h]
\caption{ Results using deterministic and stochastic Transformers for both the entire dataset and SLT subset}

\begin{center}
\begin{tabular}{| c| c || c c c c|  c c c c| }

\hline
        Data  &  Model    & \multicolumn{4}{c }{Dev}          & \multicolumn{4}{c |}{Test}         \\ 

\  & & BLEU-1 & BLEU-2 & BLEU-3 & BLEU-4 & BLEU-1 & BLEU-2 & BLEU-3 & BLEU-4 \\ 
\hline\hline
 Raw &   Stochastic     &  10.4 & 2.14 &  0.95 &  0.33 & 11.50 & 2.85 & 1.05 & 0     \\ 
 &  Deterministic     &  6.23  &	1.23  &	0.54  & 0.36 &	7.68 &  1.83 &	 0.5 &	0     \\

\hline
SLT  & Stochastic       &  21.30   &    12.26  &	8.74 &	\textbf{ 6.67} & 19.99&	 11.10&	 7.68&	 5.69   \\ 
&  Deterministic    &  18.79  &	9.69  &	 6.68 &	5.08 &  17.37 &	 8.50 &	5.35 &	 3.85      \\

\hline
\end{tabular}
\label{tab:full_vs_sample_belu}
\end{center}
\end{table*}

\subsection{Experimental setup}

This section presents our experimental results, primarily focusing on the LWTA-Transformer's application on the  SLT subset of the Elementary dataset. The suggested LWTA-Transformer version is a three-layer architecture with an embedding size of 256 and U=2 competing units.

%c+p from 2021

Following the recommendations of \cite{Camgoz20}, and \cite{NEURIPS2019_9015} we initialised the trainable parameters  using Kaiming uniform   for stochastic models \cite{he2015delving}, and  Xavier normal for deterministic models \cite{glorot2010understanding}. The Gumbel-Softmax temperature was set following the related theory in \cite{jang2017categorical}; we use a high temperature of $T = 1.00$ during training and a low $T=0.01$ during inference. The rest of the training hyperparameters were either  chosen based on the exact suggestions from the original papers of the models or optimised during our experimental investigation. The BLEU-4 score was used as the main evaluation metric to assess the quality of sign-to-text translation. Core parts of the model's implementation are modified versions of \cite{kreutzer-etal-2019-joey, Camgoz20, Voskou_2021_ICCV}.

\subsection{ Quantitative Results}

Two central inquiries targeted in our experiments are (i) the identification of the network architecture that produces the best outcomes;  and (ii) quantifying the impact of our decision to use an SLT-suitable subset rather than working with the entire Elementary23 dataset. Table \ref{tab:full_vs_sample_belu}  presents the best results that were achieved for all the subcases.

Table \ref{tab:full_vs_sample_belu} clearly demonstrates that using the Raw dataset resulted in very low BLEU scores for both the deterministic and stochastic models.   Conversely, the usage of the SLT-subset resulted in significantly improved outcomes.  The exclusion of problematic or unsuitable sentences and an appropriate data split apparently play a crucial role,  making training   less noisy and more focused on effective  examples.

In terms of architecture, we compare  two of  the best approaches available, that is, the original S2T deterministic Transformer and its stochastic counterpart, i.e. the sLWTA  Transformer. The results are summarised in Table \ref{tab:full_vs_sample_belu}. The stochastic LWTA Transformer appears to be  superior to the deterministic, achieving a BLEU-4 score of more than 1.5 units higher than the latter. The superiority holds for all ranks of BLEU scores  in both the training and validation set.

\subsubsection{Ablation study}

\paragraph{Model size   }

Contemporary NLP Transformer networks have a tendency towards large size \cite{lan2019albert,brown2020language,devlin2018bert}, reaching depths that can approach 100 layers. Conversely, SL Transformers  are commonly much smaller, often no more than 2 to 3 layers deep. Our experiments, conducted on the Elementary23-SLT dataset, aimed to determine the optimal depth for the proposed stochastic Transformer. The results, depicted in Table \ref{tab:depth}, demonstrate that a depth of 2 proves to be the optimal choice, as indicated by the highest BLEU-4 scores on both the test and validation(dev) sets. A depth of 1 delivered results that were lower but still close, while increasing the depth beyond 2 resulted in a decline in performance of around 1 to 1.5 units.
\begin{table}[h]
\caption{BLEU-4 Scores per model depth }

\begin{center}
\begin{tabular}{| c | c c |}

\hline
       Depth    & Dev          & Test        \\

\hline \hline
1       &  	6.35 &	 5.57    \\ 
2        & 6.67   & 5.69 \\
3        & 5.09   & 4.63 \\
 
\hline
\end{tabular}
\label{tab:depth}
\end{center}
\end{table}
The embedding size is another size-related hyperparameter crucial  in  Deep  NLP models. Table \ref{tab:embsize} presents a study of the effect of different embedding sizes on the performance of the LWTA-Transformer on the Elementary23 dataset. The results indicate that an embedding size of 256 provides optimal performance.  Smaller sizes appear insufficient to handle the complexity of the task, as they yielding  the much lower scores of 4.39/4.07 for size=128. On the other hand, larger sizes, such as 512, do not improve the results.
\begin{table}[h]
\begin{center}
\caption{BLEU-4 Comparison between embedding sizes}

\begin{tabular}{ |c| c c |}
\hline  
       Embedding Size     & Dev          & Test         \\

\hline \hline
128       &  4.39 &	 4.07   \\ 
256        &  	6.67 &	 5.69    \\ 
512        &  5.32   & 5.27 \\

\hline
\end{tabular}
\label{tab:embsize}
\end{center}
\end{table}
\paragraph{The effect of Competing Units per Block }
As previously discussed, the results of our experiments on the sLWTA Sign Language Transformer (Table \ref{tab:full_vs_sample_belu}) render it a superior solution for the Greek SL translation task compared to the deterministic model. A central aspect of this network is the  use of the sophisticated LWTA layer, as opposed to a typical activation function such as Relu. The size of the competition blocks $U$ is the main tunable hyperparameter of this technique. Through an examination of the commonly used sizes, presented in Table \ref{tab:units}, we  concluded that the most suitable choice for our case is $U=2$, as suggested in \cite{panousis2019nonparametric}. Larger sizes of $U=4$ and $U=8$ resulted in decreases of 0.22 and 0.76 BLEU-4 units, respectively; this is likely due to the high sparsity of the  representations obtained from the LWTA blocks.

\begin{table}[ht]
\begin{center}
\caption{  The effect of LWTA block size U}

\begin{tabular}{ |c| c c| }

\hline 
       Competing Units (U)       & Dev          & Test         \\

\hline \hline
2       &  	6.67 &	 5.69    \\ 
4        & 5.41   & 5.47 \\
8        & 5.23   & 4.93 \\

\hline
\end{tabular}
\label{tab:units}
\end{center}
\end{table}

\subsubsection{Discussion and Benchmarking}

We now proceed with a direct evaluation of the translation accuracy attained on the Elementary23 dataset, and compare it to the results reported on other benchmark datasets. Table \ref{tab:becnmarks} summarises  the achieved BLEU-4 scores for each case, covering both the main benchmarks and a curated selection of significant yet less directly comparable supplementary non-European datasets.

\begin{table}[h]
\begin{center}
\caption{Benchmarking BLEU-4 scores }

\begin{tabular}{|l|ll|l|}
\hline
Dataset       &                 Dev & Test   \\
\hline
\hline
Elementary23-SLT &       6.67 & 5.69   \\
\hline
Phoenix2014T \cite{Voskou_2021_ICCV} &     23.23  & 23.65   \\
SWISSTXT-NEWS \cite{camgoz2021content4all}         &   0.46  &   0.41    \\
VRT-NEWS \cite{camgoz2021content4all}               &  0.45   &  0.36    \\
\hline
OpenASL \cite{shi2022open} &    6.57 & 6.72    \\
CSL-Daily \cite{zhou2021improving}               &  20.80   &  21.34     \\
\hline
\end{tabular}
\label{tab:becnmarks}
\end{center}
\end{table}

As indicated in this table, researchers have reported BLEU-4 scores reaching as high as 23.23/23.65 for the validation and test sets of Phoenix2014T, validating its standing as the highest-performing dataset. However, as previously noted, this dataset does come with limitations such as a restricted vocabulary, a narrowly focused topic, and a stringent structure. Analogous results emerge from applications on the CSL-Daily dataset. More specifically, researchers report impressive scores of 20.80/21.34 on this popular Chinese-SL dataset, which also bears similar constraints on vocabulary and content to Phoenix2014T. While these attributes enhance BLEU-4 performance, they diminish the applicability of the developed SLT models for real-world users.
In contrast, the objective of our work is to amplify the effectiveness of end-to-end SLT systems in genuine usage scenarios. These considerations make Phoenix2014T an imperfect comparison to the Elementary23 dataset, which boasts a more realistic design.

Conversely, models trained on SWISSTXT-NEWS and VRT-NEWS  exhibit weak performance, with BLEU-4 scores $<1$. The authors report scores of 0.46/0.41 and 0.45/0.36, respectively, which are considerably lower than the 6.67/5.69  achieved in our proposed subset. Unfortunately, with such poor scores, these datasets cannot provide any practical value, nor can they be regarded as a reliable benchmark for future SLT models; these facts are despite their extensive topic coverage and vast vocabulary size.

These contrasting outcomes underscore the value of the new Elementary23 dataset. Although the attained  BLEU-4 scores, around 6 units, are  low compared to state-of-the-art NLP models trained on extensive text corpora, they demonstrate tangible translation capabilities.   Therefore, unlike the  previously mentioned SL datasets, our data combine measurable results with a comprehensive and realistic thematic spectrum. These qualities give our proposed dataset particular importance as a benchmark dataset for future SLT research. Finally, the more modern OpenASL is the only case that seems to align with Elementary23, combining respectable results with high-quality content.

\subsection{Qualitative Results}

The quality of the automatic translations produced by our models  varies; this is shown in Table \ref{tab:qual_examples}, where three representative examples are presented. In some cases, such as the first example, the results are impressively accurate, with only minor numerical or grammatical errors. The next  case belongs to a second category in which the context is partially  captured, but the syntax deviates from the target. The final example represents the third group, where the model completely fails to detect any of the signed signals.

\begin{table}[h]
\footnotesize
\begin{center}
\caption{Reference (R), Prediction (P), Translated Reference (Rt), Translated Prediction (Pt)}

\begin{tabular}{|l|l|}

\hline 
\# & Translation Reference and Prediction \\
\hline
\hline

1 & \textbf{R:} 
\selectlanguage{greek}
έχει 10 νομίσματα πόσα είναι τα χρήματά του συνολικά
\selectlanguage{english} \\
& \textbf{P:} 
\selectlanguage{greek}
έχει 4 νομίσματα πόσα είναι τα χρήματά του συνολικά 
\selectlanguage{english} \\

& \textbf{Rt:} 
\selectlanguage{english}
he has 10 coins how much is his money in total 
\selectlanguage{english} \\
& \textbf{Pt:} 
\selectlanguage{english}
he has 4 coins how much is his money in total
\selectlanguage{english} \\

% ( the wind blows weak to moderate in the south)\\
\hline
2 & \textbf{R:}  
\selectlanguage{greek}
συμπληρώνω τους αριθμούς που λείπουν στους πίνακες
\selectlanguage{english} \\
&\textbf{P:} 
\selectlanguage{greek}
 υπολογίζω και γράφω τους αριθμούς που λείπουν 
\selectlanguage{english} \\

% ( on friday there are lots of clouds and showers in some areas)\\

& \textbf{Rt:} 
\selectlanguage{english}
I fill in the missing numbers in the tables
\selectlanguage{english} \\
&\textbf{Pt:} 
\selectlanguage{english}
I calculate and write  the missing numbers
\selectlanguage{english} \\

\hline
3 & \textbf{R:} 
\selectlanguage{greek}
στο σχολείο μαθαίνουμε καινούρια πράγματα 
\selectlanguage{english} \\
& \textbf{P:} 
\selectlanguage{greek}
  το περιβάλλον μου 
\selectlanguage{english} \\

% ( in the west and northwest there are some showers in the west)\\

& \textbf{Rt:} 
\selectlanguage{english}
at school we learn new things
\selectlanguage{english} \\&\textbf{Pt:} 
\selectlanguage{english}
my environment
\selectlanguage{english} \\

\hline 
\end{tabular}
\label{tab:qual_examples}
\end{center}
\end{table}
\section{Conclusions}

This paper presents a novel dataset of Greek Sign Language, characterized by high technical quality and diverse vocabulary. Using a simple yet effective selection procedure, we selected an SLT-suitable subset that adheres to established formatting standards and retains the desired features. By utilizing variants of the state-of-the-art LWTA Transformer, we achieved BLEU-4 scores ten times higher than those reported on directly comparable datasets. 

As such, our proposed Elementary23 dataset offers a balance of quality and feasibility. Elementary23 can serve as a viable alternative to  popular yet monotonous dataset like Phoenix2014T  and similar, and the verbally rich but unattainable such as SWISSTXT-NEWS and VRT-NEWS options. Future work could involve applying even more sophisticated SLT models, exploring data-oriented techniques such as active or meta-learning, and further investigating the use of the Elementary23 dataset for tasks such as text-to-sign SL production.

\section*{Acknowledgement}

This research was supported by the  European Union’s Horizon 2020 research and innovation program, under grant agreement No 872139, project aiD.
 
{\small
\bibliographystyle{ieee_fullname}
\bibliography{elementary23}
}

\end{document}